\documentclass[a4paper]{article}

\usepackage{amsmath,graphicx,mathtools,amssymb,setspace,lipsum,cuted,multicol,multirow,makecell,hyperref,cite,INTERSPEECH2019}

\title{Incorporating Symbolic Sequential Modeling for Speech Enhancement}
\name{Chien-Feng Liao$^1$,  Yu Tsao$^1$, Xugang Lu$^2$, Hisashi Kawai$^2$}
\address{
  $^1$Research Center for Information Technology Innovation, Academic Sinica, Taiwan\\
  $^2$National Institute of Information and Communications Technology, Japan}
\email{r06946002@ntu.edu.tw, yu.tsao@citi.sinica.edu.tw, \{xugang.lu,hisashi.kawai\}@nict.go.jp}

\begin{document}

\maketitle
\begin{abstract}
In a noisy environment, a lossy speech signal can be automatically restored by a listener if he/she knows the language well. That is, with the built-in knowledge of a ``language model'', a listener may effectively suppress noise interference and retrieve the target speech signals. Accordingly, we argue that familiarity with the underlying linguistic content of spoken utterances benefits speech enhancement (SE) in noisy environments. In this study, in addition to the conventional modeling for learning the acoustic noisy-clean speech mapping, an abstract symbolic sequential modeling is incorporated into the SE framework. This symbolic sequential modeling can be regarded as a ``linguistic constraint'' in learning the acoustic noisy-clean speech mapping function. In this study, the symbolic sequences for acoustic signals are obtained as discrete representations with a Vector Quantized Variational Autoencoder algorithm. The obtained symbols are able to capture high-level phoneme-like content from speech signals. The experimental results demonstrate that the proposed framework can obtain notable performance improvement in terms of perceptual evaluation of speech quality (PESQ) and short-time objective intelligibility (STOI) on the TIMIT dataset.

\end{abstract}
\noindent\textbf{Index Terms}: Speech enhancement, deep learning, symbolic representation, multi-head attention

\section{Introduction}
\label{sec:intro}
Speech enhancement (SE) has been commonly used as a front-end module in speech-related applications, such as robust automatic speech recognition (ASR) \cite{li2014overview, li2015robust, wang2016joint}, automatic speaker recognition, and assistive listening devices \cite{loizou2007speech, lai2017deep, wang2017deep}. Recently, deep learning (DL)-based SE models have also been proposed and extensively investigated \cite{lu2013speech, xu2015regression, chen2015speech, kolbk2017speech, wang2018supervised}. The main idea in these DL-based SE models is to learn the complex mapping functions between noisy speech and clean speech. In most studies, the mapping functions are learned based on a large quantity of well-prepared noisy-clean speech pairs in the acoustic domain without considering the underlying linguistic structure. 

In a noisy environment, audiences can automatically restore a noise-masked speech based on their knowledge of a ``language model'', and the restoring ability depends on the effectiveness of this internal ``language model''. For example, in noisy environments, great effort is required for non-native listeners \cite{borghini2018listening}. These studies indicate that the linguistic-related information is helpful to retrieve target speech signals from the noisy ones. Accordingly, it is argued in this study that it is beneficial to incorporate text information (phonemes or words) into an SE system for improved performance.

In \cite{kinoshita2015text}, oracle transcription is used to extract time-aligned text features as auxiliary input to the DNN model. Even though this can be formulated as a text-to-speech application, it is not practical under SE scenarios to assume to have ground-truth transcription. Several studies incorporate recognition results or outputs from acoustic models. In \cite{mimura2015deep}, a phone-class feature is augmented to standard acoustic features as input for de-reverberation. In \cite{chen2015speech}, an ASR and an SE system are trained iteratively, where each system's input depend on the other's output. In \cite{wang2016phoneme, chazan2016phoneme}, a set of DNNs were trained as enhancement models, one for each specific phoneme. During inference time, an ASR or a phoneme classifier was used to determine which DNN to use. Even though promising results have been obtained, these approaches have major drawbacks. First, the recognition model is not jointly trained and thus optimization cannot be achieved for both systems. If the ASR system is incorrect, errors will be propagated to the downstream SE system. Secondly, heavily equip SE with an ASR system may be undesirable because SE is commonly used as a preprocessor. To overcome these obstacles, \cite{chazan2017speech} proposed learning a Deep Mixture of Experts (DMoE) network where the experts are DNNs, whose outputs are combined by a gating DNN. The gating DNN is trained to assign a combination weight to each expert. This results in splitting the acoustic space into sub-areas in an unsupervised manner, which is similar to our proposed method.

van den Oord et al. \cite{van2017neural} recently proposed the Vector Quantized Variational Autoencoder (VQ-VAE), in which the stochastic continuous latent variables from the original VAE are replaced with deterministic discrete latent variables. It maintains a set of prototype vectors, i.e., a predefined size of learnable codebook. During forward pass, feature vectors produced by the encoder are replaced with their nearest-neighbor in the codebook. Although this quantization component acts as an information bottleneck and can regularize the power of the encoder, the discrete latent variables are more interpretable and tend to learn higher level representations, which can naturally correspond to phoneme-like features for given speech signal inputs. In \cite{chorowski2019unsupervised}, a comprehensive study of VQ-VAE applied to speech data was carried out, and it was demonstrated that VQ-VAE achieves better interpretability and information separation (such as disentangling speaker characteristics) than VAEs and AEs. Furthermore, the extracted representation allowed for accurate mapping into phonemes and achieved competitive performance on an unsupervised acoustic unit discovery task. Overall, the characteristics of the VQ-VAE make it a suitable component to reinforce an SE system with high-level linguistic information. 

\begin{figure}[ht]
\includegraphics[width=0.85\linewidth, keepaspectratio]{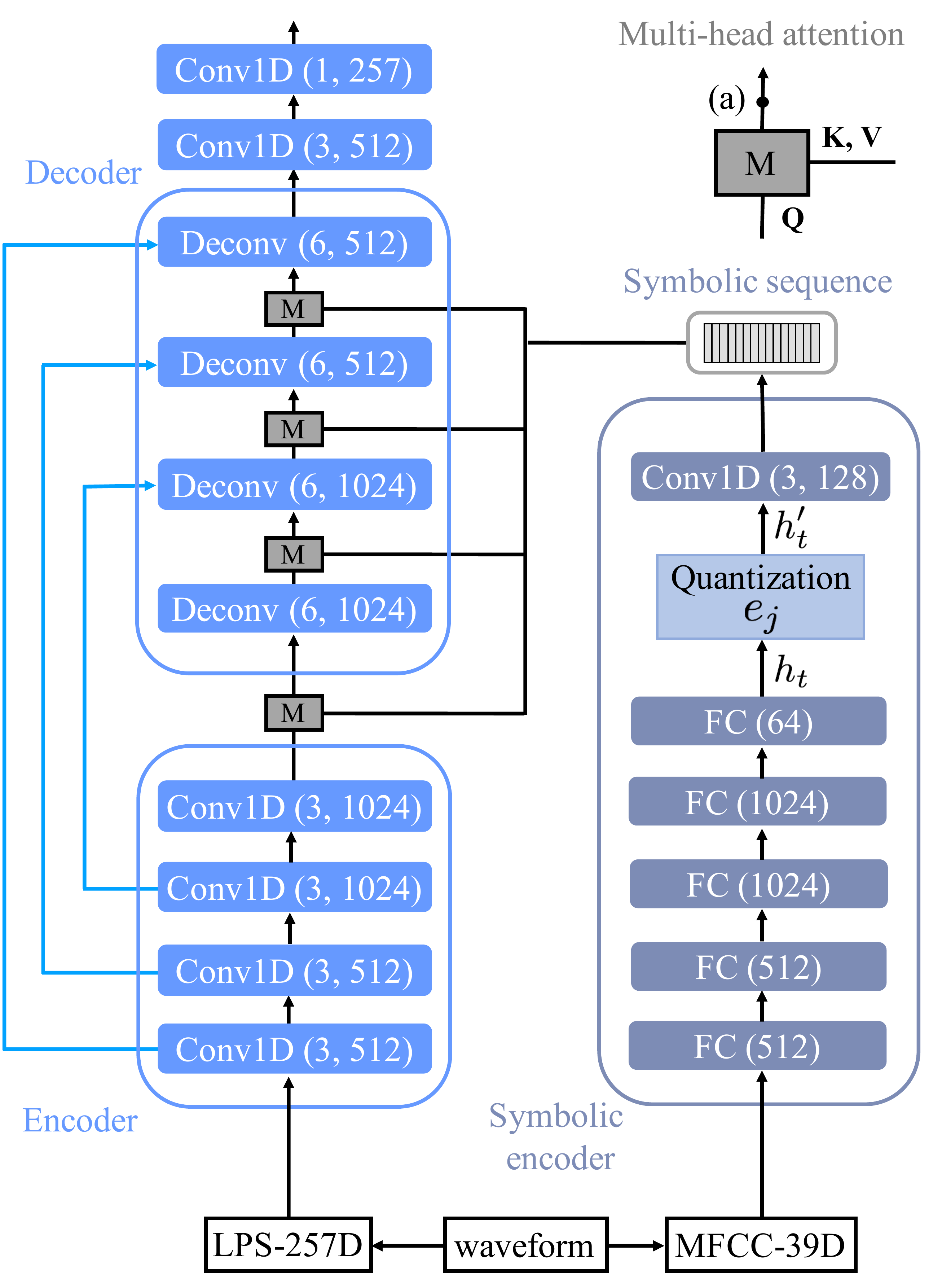}
 \centering
  \caption{Proposed system consisting of a U-Net architecture, a symbolic encoder, and an attention mechanism. Conv1Ds and Deconvs are in the format (filterWidth, outputChannels), and the down-sample$\backslash$up-sample rates are both 2. FC (outputChannels) denotes the fully connected layer.}
  \label{fig:system}
\vspace{-0.15in}
\end{figure}

In this study, an SE system with U-Net architecture \cite{ronneberger2015u, pascual2017segan, michelsanti2017conditional, stoller2018wave} is proposed. Moreover, a ``symbolic encoder'' is developed, consisting of DNNs and the vector quantization mechanism in VQ-VAE. The extracted symbolic sequence is then connected to the U-Net via multi-head attention mechanism \cite{vaswani2017attention}. Thereby, the two components can be jointly trained without the need of any supervised transcription or explicit constraints. The results demonstrate a notable improvement in terms of objective measures including perceptual evaluation of speech quality (PESQ) \cite{recommendation2001perceptual} and short-time objective intelligibility (STOI) \cite{taal2011algorithm}.

The rest of the paper is organized as follows. In Section \ref{sec:symb_enc}, the proposed approach is detailed, including each components of the system and the objective functions. The experiment settings and results are presented in Section \ref{sec:exp}. Finally, Section \ref{sec:conclusion} concludes the paper.

\section{System architecture}
\label{sec:symb_enc}
A paired training dataset $\{x_i,y_i\}^N_{i=1}$, where $x_i$ is the input noisy speech and $y_i$ is the target clean speech. The proposed system is shown in Figure \ref{fig:system}. It consists of the following parts: an encoder network $E_{se}(x)$ consisting of convolutional layers that extracts the feature sequence; another encoder network called \textbf{symbolic encoder} $E_{symb}(x)$ consists of fully connected layers and extracts the \textbf{symbolic sequence} by vector quantization. Multi-head attention function and skip-connection are used to connect the two encoder outputs with the decoder $Dec(E_{se}(x), E_{symb}(x))$. All components are jointly trained using mean-squared-error (MSE) loss function between the clean speech and the enhanced speech:
\begin{equation}
    \mathcal{L}_{mse} = 
    \frac{1}{N} \sum_{i=1}^{N} || Dec(E_{se}(x_i), E_{symb}(x_i)) - y_i ||^2_2
  \label{loss_mse}
\end{equation}
The quantization mechanism and the multi-head attention mechanism will now be briefly explained; for more detailed information readers may refer to \cite{van2017neural} and \cite{vaswani2017attention}, respectively.

\begin{table*}[ht]
\caption{Average PESQ, and STOI scores for evaluating baseline models and the proposed method on the test set under three unseen noise environments at five SNR levels and the average scores across all SNRs. The unprocessed test set is denoted by \textbf{Noisy}. Size of the symbolic book is shown in the parenthesis. The highest scores per metric are highlighted with bold text, excluding \textbf{Oracle}.}
\centering
\begin{tabular}{ |c || c|c || c|c || c|c || c|c || c|c || } 
 \toprule
    \multicolumn{1}{ |c||}{} & 
        \multicolumn{2}{c||}{Noisy}&
            \multicolumn{2}{c||}{U-Net}&
                \multicolumn{2}{c||}{U-Net-MOL}&
                    \multicolumn{2}{c||}{Proposed (64)}&
                        \multicolumn{2}{c||}{Oracle} \\
 \cline{2-11}

    SNR &PESQ  &STOI &PESQ  &STOI &PESQ  &STOI &PESQ  &STOI &PESQ  &STOI \\
\hline
-6   & 1.213 & 0.532 & 1.685 & 0.602 & 1.800 & 0.619 & \textbf{1.828} & \textbf{0.624} & 1.961 &  0.703  \\
-3   & 1.353 & 0.598 & 1.880 & 0.669 & 1.974 & 0.681 & \textbf{2.045} & \textbf{0.693} & 2.140 &  0.741  \\
0    & 1.517 & 0.669 & 2.071 & 0.725 & 2.140 & 0.736 & \textbf{2.240} & \textbf{0.750} & 2.306 &  0.776  \\
3    & 1.702 & 0.739 & 2.237 & 0.770 & 2.290 & 0.779 & \textbf{2.416} & \textbf{0.794} & 2.456 &  0.806  \\
6    & 1.902 & 0.823 & 2.387 & 0.805 & 2.424 & 0.813 & \textbf{2.581} & \textbf{0.830} & 2.592 &  0.831  \\
Avg. & 1.537 & 0.669 & 2.052 & 0.714 & 2.126 & 0.725 & \textbf{2.222} & \textbf{0.738} & 2.291 &  0.771  \\
\bottomrule

\end{tabular}
 \label{table1}
\end{table*}

\subsection{Symbolic Encoder}
The symbolic encoder reads a sequence of acoustic features as input. Here, mel-frequency cepstral coefficients (MFCCs) are used, as suggested in \cite{chorowski2019unsupervised}. A sequence of hidden vectors $\{h_t\in{R^D},t=1,...,T\}$ is extracted by the fully connected layers, where $D$ is the dimensionality and $T$ denotes the sequence length. A \textbf{symbolic book} that contains a set of prototype vectors $\{e_j\in{R^D} ,j=1,...,M\}$ is maintained, where $M$ is the size of the book. The hidden vectors $h_t$ will be replaced by the nearest prototype vector in the symbolic book. That is, $h'_t = e_k$, where $ k=argmin_j\|h_t - e_j\|^2_2$. During the training phase, the prototypes in the symbolic book are updated as a function of exponential moving averages of $h$. This method is presented in the original paper as an alternative way to update the book, and has the advantage of faster training speed than using an auxiliary loss. To prevent the symbolic encoder diverge in $h$ with unbounded value, \cite{van2017neural} also uses the ``commitment loss'' to encourage the symbolic encoder to produce vectors lying close to the prototypes. Overall, the full system is optimized with two loss terms: the MSE between the enhanced acoustic features and the clean target features, and the commitment loss:
\begin{equation}
    \mathcal{L}_{total} = \mathcal{L}_{mse} + \lambda \| h_t - sg(e_k)\|^2_2
  \label{loss_total}
\end{equation}
where $\lambda$ is a hyperparameter that controls the importance of the commitment loss and $sg(.)$ denotes the stop-gradient operation. It should be noted here that the gradient of the loss can be backpropagated to the symbolic encoder using the straight-through estimator presented in \cite{bengio2013estimating}.
 
\subsection{Multi-head Attention}
Multi-head attention (MHA) was first proposed in the transformer architecture \cite{vaswani2017attention} for machine translation, and recently explored in various speech-related tasks including end-to-end ASR \cite{chiu2018state} and text-to-speech system \cite{wang2018style}. MHA extends the conventional attention mechanism to have multiple heads, where each head generates a different attention weight vector. This allows the decoder to jointly retrieve information from different representation subspaces at different positions, which facilitates focusing on the various structures of the symbolic sequence. The input argument consists of queries $Q$, keys $K$, and values $V$, i.e., $Attn(Q,K,V)$. In this study, MHA is used before each layer in the decoder. Every time-step of the decoder output acts as an query to attend on the symbolic sequence. The output of MHA will be concatenated with the skip-connection and fed to the proceeding decoder layer together. Formally, we have the symbolic sequence $h$ and the skip-connection from the encoder at each layer $\{s^{(l)}, l=1,...,L\}$. The output of each layer in the decoder is the following:
 \begin{align*}
 \label{MHA}
    & d^{(l)} = Deconv(Concat(s^{(L-l+1)},Attn(d^{(l-1)},h',h')))
 \end{align*}
where $l=1,...,L$ is the depth of the decoder layer and $d^{(0)}$ is the encoder output. 

\subsection{Model Details}
\label{ssec:model}
The symbolic encoder consists of four fully connected layers, each followed by a ReLU activation function and a dropout layer \cite{srivastava2014dropout} with a drop rate of 0.2. A linear projection layer then maps the hidden vectors $h_t$ to $D = 64$ dimensions in order to perform quantization. After the quantization, an one-dimensional (1-D) convolutional layer is used to give the symbolic sequence contextual information. Four heads are used in MHA, leading to point (a) in Figure \ref{fig:system}, with a dimensionality of $4\times128$. As in the original transformer, the positional encodings are also added to the inputs of the MHA, providing some information about the position of the tokens in the sequence. Queries and keys are first passed through a linear projection layer with 256 nodes before being divided into multiple heads. For the encoder, the frequency axis is treated as channel; thus, 1-D convolutional layers are used. The sequence length is down-sampled at each layer using a stride of 2 instead of pooling layers. The decoder is a mirrored version of the encoder with deconvolutional layers and larger kernel width. LeakyReLU is used as activation function in both the encoder and the decoder. Finally, the decoder output is projected back to frequency dimension using 1-D convolution with a kernel width of 1.

\begin{figure*}[t]
\includegraphics[width=0.78\linewidth, keepaspectratio]{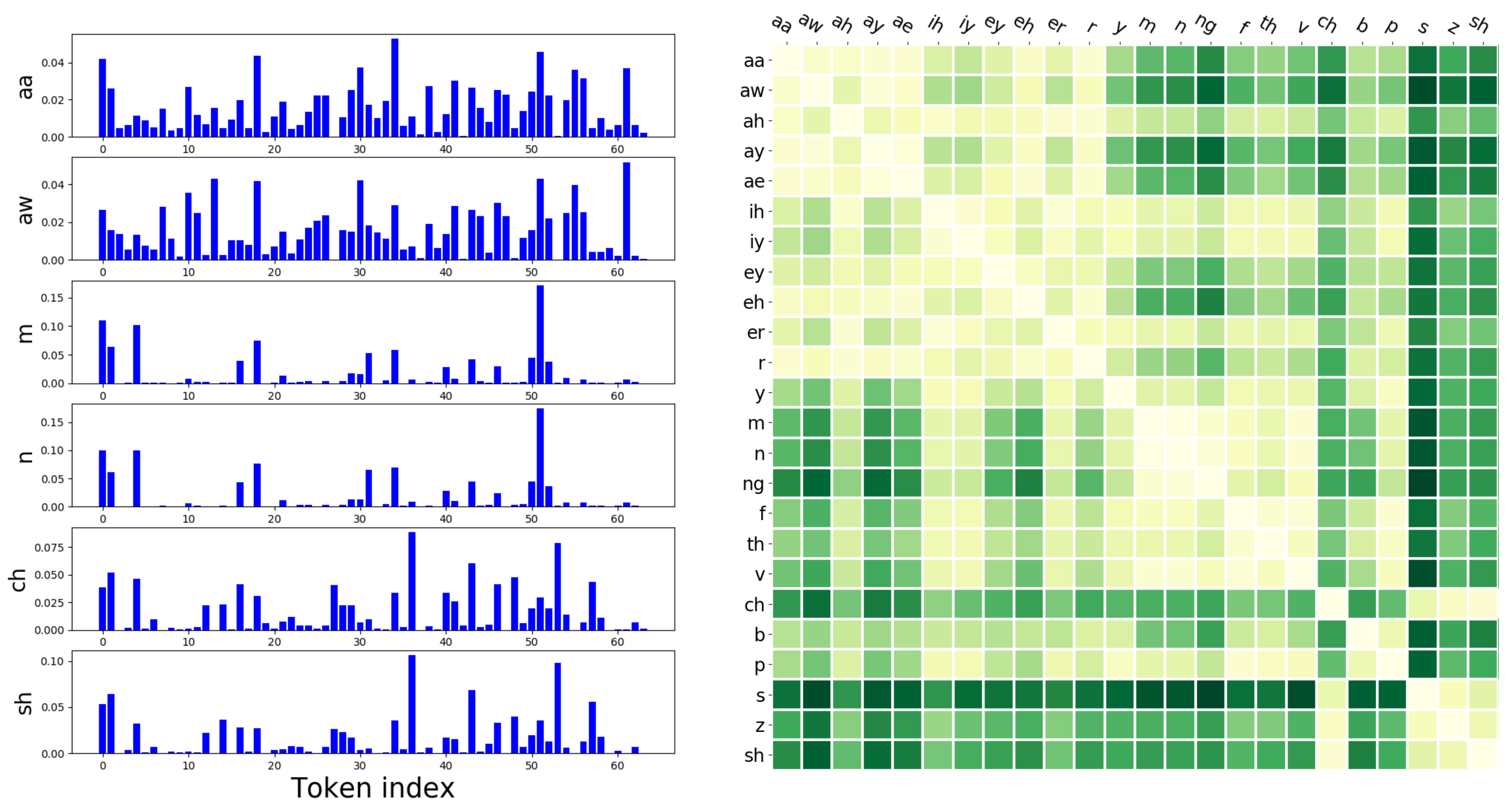}
\centering
  \caption{Left: Histogram: each bin represents the token index, and the value shows how many times this token was chosen, given the corresponding phoneme. Right: The element on location (i,j) represents JS-divergence between the histogram from the i-th phoneme and the histogram from the j-th phoneme. Darker color implies larger divergence. Some phonemes were omitted owing to space limitations.}
  \label{fig:visualize}
\end{figure*}

\section{Experiments}
\label{sec:exp}
The experiments were conducted on the TIMIT database \cite{garofolo1993timit}. A total of 3696 utterances from the TIMIT training set (excluding SA files) were randomly sampled and corrupted with 100 noise types from \cite{hu100} at six SNR levels, i.e., 20dB, 15dB, 10dB, 5dB, 0dB, and -5dB, to obtain 40-hour multi-condition training set, consisting of pairs of clean and noisy speech utterances. Another 100 utterances were randomly sampled to construct the validation set. They are mixed with cafeteria babble noise at 4 SNR levels (-4 dB, 0 dB, 4 dB, and 8 dB), which is unseen from the training set. The 192 utterances from the core test set of the TIMIT database were used to construct the test set for each combination of noise types and SNR levels. To evaluate the system on unseen noise types, three other noise types, namely Buccaneer1, Destroyer engine, and HF channel from the NOISEX-92 corpus \cite{varga1993assessment}, were adopted. In the following experiments, the SE algorithm will be evaluated in terms of speech quality and speech intelligibility. Therefore, PESQ and STOI, respectively, will be used to evaluate the enhanced speech, respectively. Higher scores represent better performance.

\subsection{Implementation}
The sampling rate of the speech data was 16 kHz. For the encoder input, time-frequency (T-F) features were extracted  using a 512-point short time Fourier transform (STFT) with a hamming window size of 32 ms and a hop size of 16 ms, resulting in feature vectors consisting of 257-point STFT log-power spectra (LPS). For the symbolic encoder, standard 13 MFCC features (extracted at a rate identical to that for the LPS features) were used and concatenated with their temporal first and second derivatives. MFCCs are often used in speech recognition because they are pitch invariant and slightly robust to noise. A better quantization behavior was observed using MFCC compared to LPS in the preliminary experiments. The input was a segment of 64 frames (approximately 1 s), and was normalized by mean and standard-deviation before being fed to the system. Finally, the decoder outputs were synthesized back to the waveform signal via inverse Fourier transform and an overlap-add method. The phases of the noisy signals were used for the inverse Fourier transform. All models were trained on minibatches of 32. The Adam optimizer \cite{kingma2014adam} was used with learning rate $lr=0.0001$, $\beta_1=0.5$, and $\beta_2=0.9$. The weight of the commitment loss $\lambda$ was set to 0.2, which is close to the original setting in VQ-VAE, and it did not have significant impact on performance. Early stopping was performed based on the validation set to prevent overfitting. 

\subsection{Baseline model}
We constructed the baseline model by excluding the symbolic encoder component, i.e., the left part of Figure \ref{fig:system} without MHA. This model is denoted by \textbf{U-Net}. Subsequently, the multi-objective learning method proposed in \cite{xu2017multi} was adopted in the baseline model. The input of the \textbf{U-Net} was augmented by MFCC features, and an additional objective was added to $\mathcal{L}_{total}$ during training to predict clean MFCCs. This baseline is denoted by \textbf{U-Net-MOL}. Finally, the benefit of using real text information as in \cite{kinoshita2015text} should be demonstrated. The phoneme level transcriptions provided by the TIMIT corpus were used to obtain frame-wise phoneme labels. The input MFCCs of the symbolic encoder were then replaced by the phoneme embeddings (embeddings are jointly learned). Quantization was discarded because the real phonetic information was provided. This is considered as an oracle model, as it takes correct transcriptions as input. This system will be called \textbf{Oracle}.

\begin{table}[t!]
  \caption{Average PESQ and STOI performance on the validation set for different size of the symbolic book.}
  \label{tab:numtoken}
  \centering
  \begin{tabular}{ c c c }
    \toprule
    \textbf{Book size \textit{M}} & \textbf{PESQ} & \textbf{STOI} \\
    \midrule
    39      & 2.061             & 0.711                 \\
    64      & \textbf{2.108}    & \textbf{0.713}                 \\
    128     & 2.027             & 0.712        \\
    256     & 2.041             & 0.711                 \\
    \bottomrule
    \vspace{-0.3in}
  \end{tabular}
\end{table}
\subsection{Results}
Table \ref{table1} presents the results of the average PESQ, SSNR, and STOI scores on the test set for different systems. ``Noisy'' denotes unprocessed noisy speech, and the proposed model is shown with the symbolic book size of 64 as a representative. From this table, it can be observed that \textbf{Oracle} performed the best, as expected. This also confirmed the hypothesis that, given correct text information, the SE system can be more robust to noisy environments. Furthermore, the proposed model outperformed \textbf{U-Net} and \textbf{U-Net-MOL} at every SNR levels. It should be noted here that the system had fewer trainable parameters compared to the baselines, as MHA reduces the dimension to $4\times128$, as mentioned in Section \ref{ssec:model}. Thus, the improvement was not due to model complexity. Table \ref{tab:numtoken} shows the de-noise ability of the proposed method with different size symbolic book. It can be seen that performance peaked for a size of 64. During the experiments, it was also observed that the symbolic book suffered from the ``index collapse'' problem \cite{kaiser2018fast} (some tokens are not activated through out training) for sizes larger than 256, implying that 256 tokens are sufficient for exploring the acoustic units, whereas adding more will be of no benefit.

\subsection{Interpretation of symbolic sequence}
An advantage of the discrete representation learned by the VQ-VAE is the interpretability of individual tokens in the symbolic book. Here, a visualization method was developed to connect input acoustic features to the activated token. Figure \ref{fig:visualize} (left) shows histograms corresponding to phoneme classes (39-way). More specifically, noisy speech from the test set were passed through the symbolic encoder to obtain the symbolic sequences. Given the frame-wise phoneme labels, a histogram for individual phoneme class can then be plotted. Each bin represents the token index, and the value shows how many times this token was chosen, given the frame that belongs to the corresponding phoneme. The histograms were normalized to become probability distribution functions (PDFs), i.e., the summation equals 1. Here, it can be seen that phonemes with similar pronunciation also have similar distribution in the histograms. For example, the phonemes in each of the pairs (\textit{aa}, \textit{aw}), (\textit{m}, \textit{n}), and (\textit{ch}, \textit{sh}) have similar distributions, whereas phonemes in different pairs have different distributions.

For a complete understanding of the relations within the phoneme set, the Jensen-Shannon divergence between the phonemes was measured. Figure \ref{fig:visualize} (right) shows a heat map. Each element represents the distance between two PDFs, and darker color corresponds to larger distance. As JS-divergence is symmetric, the heat map is also a symmetric matrix. Some squares in light color are located on the diagonal, which implies that phonemes with similar pronunciation are clustered together, e.g., vowels have lighter colors with each other, and are completely separated from fricatives. The heat map greatly facilitates the visualization of the relationship between phonemes. For instance, it shows that \textit{ch} is very close to \textit{s}, \textit{z}, and \textit{sh}. In conclusion, the symbolic encoder was demonstrated to be reactive to phonetic content. It was observed that some of the phonemes that are pronounced differently lie near each other. The obvious explanation is that the noise affected the input MFCCs, thus confusing the symbolic encoder. One possible solution is to constrain explicitly the symbolic encoder so that it may become noise-invariant by adding a discriminator and using adversarial training as in \cite{liao2018noise}. This is left as future work.

\section{Conclusion and future work}
\label{sec:conclusion}
A novel approach for incorporating phonetic content into a SE system was proposed, without the need for a recognition system or any transcriptions during training. The symbolic encoder used the vector quantization method proposed in VQ-VAE to extract discrete representations. Consequently, the symbolic encoder learned to divide the input MFCCs into acoustic units automatically, and achieved notable performance improvement compared to the baseline systems. The representations were further interpreted by visualizing the symbolic encoder behavior, and it was confirmed that it was phoneme-sensitive. In future studies, the effect of different noise types on the symbolic encoder will be investigated, and noise-invariant training will be performed to extract purer symbolic sequence. Furthermore, an explicit language model constraint based on the learned symbolics may be even more useful to the SE system. 

\bibliographystyle{IEEEtran}
\bibliography{mybib}


\end{document}